\documentclass{article}

\usepackage{microtype}
\usepackage{graphicx}
\usepackage{booktabs} % for professional tables

\usepackage{hyperref}
\usepackage{multirow}
\usepackage{adjustbox}
\usepackage{pifont}
\usepackage{subcaption}
\usepackage{verbatim}
\usepackage{listings}
\usepackage{color}
\definecolor{codegreen}{rgb}{0,0.6,0}
\definecolor{codegray}{rgb}{0.5,0.5,0.5}
\definecolor{codepurple}{rgb}{0.58,0,0.82}
\definecolor{backcolour}{rgb}{0.95,0.95,0.92}

\lstdefinestyle{mystyle}{
backgroundcolor=\color{backcolour},   
commentstyle=\color{codegreen},
keywordstyle=\color{magenta},
numberstyle=\tiny\color{codegray},
stringstyle=\color{codepurple},
basicstyle=\ttfamily\footnotesize,
breakatwhitespace=false,         
breaklines=true,                 
captionpos=b,                    
keepspaces=true,                 
numbers=left,                    
numbersep=5pt,                  
showspaces=false,                
showstringspaces=false,
showtabs=false,                  
tabsize=2
}

\usepackage[accepted]{icml2024}

\usepackage{amsmath}
\usepackage{amssymb}
\usepackage{mathtools}
\usepackage{amsthm}

\usepackage{xcolor}

\usepackage[capitalize,noabbrev]{cleveref}

\theoremstyle{plain}

\theoremstyle{definition}

\theoremstyle{remark}

\usepackage[textsize=tiny]{todonotes}

\icmltitlerunning{Rethinking Post-Hoc Search-Based Neural Approaches for Solving Large-Scale TSPs}

\begin{document}

\twocolumn[
\icmltitle{Position: Rethinking Post-Hoc Search-Based Neural Approaches for Solving Large-Scale Traveling Salesman Problems}

\icmlsetsymbol{intern}{$\dag$}
\begin{icmlauthorlist}
	\icmlauthor{Yifan Xia}{nju,intern}
	\icmlauthor{Xianliang Yang}{msra}
	\icmlauthor{Zichuan Liu}{nju,intern}
	\icmlauthor{Zhihao Liu}{cas,intern}
	\icmlauthor{Lei Song}{msra}
	\icmlauthor{Jiang Bian}{msra}
\end{icmlauthorlist}

\icmlaffiliation{nju}{Nanjing University, Nanjing, China}
\icmlaffiliation{cas}{Institute of Automation, Chinese Academy of Sciences, Beijing, China}
\icmlaffiliation{msra}{Microsoft Research Asia, Beijing, China}

\icmlcorrespondingauthor{Jiang Bian}{jiang.bian@microsoft.com}

% You may provide any keywords that you
% find helpful for describing your paper; these are used to populate
% the "keywords" metadata in the PDF but will not be shown in the document
\icmlkeywords{Machine Learning in Operations Research, Traveling Salesman Problem, Large-Scale Combinatorial Problems, Heatmap-Guided Monte Carlo Tree Search}

\vskip 0.3in
]

% this must go after the closing bracket ] following \twocolumn[ ...

% This command actually creates the footnote in the first column
% listing the affiliations and the copyright notice.
% The command takes one argument, which is text to display at the start of the footnote.
% The \icmlEqualContribution command is standard text for equal contribution.
% Remove it (just {}) if you do not need this facility.

\printAffiliationsAndNotice{\icmlIntern}  % leave blank if no need to mention equal contribution
% \printAffiliationsAndNotice{\icmlEqualContribution} % otherwise use the standard text.

\begin{abstract}
	Recent advancements in solving large-scale traveling salesman problems (TSP) utilize the heatmap-guided Monte Carlo tree search (MCTS) paradigm, where machine learning (ML) models generate heatmaps, indicating the probability distribution of each edge being part of the optimal solution, to guide MCTS in solution finding. However, our theoretical and experimental analysis raises doubts about the effectiveness of ML-based heatmap generation. In support of this, we demonstrate that a simple baseline method can outperform complex ML approaches in heatmap generation. Furthermore, we question the practical value of the heatmap-guided MCTS paradigm. To substantiate this, our findings show its inferiority to the LKH-3 heuristic despite the paradigm's reliance on problem-specific, hand-crafted strategies. For the future, we suggest research directions focused on developing more theoretically sound heatmap generation methods and exploring autonomous, generalizable ML approaches for combinatorial problems. The code is available for review: \url{https://github.com/xyfffff/rethink_mcts_for_tsp}.
\end{abstract}
\section{Introduction}
\begin{figure}[ht]
	\vskip 0.2in
	\begin{center}
		\centerline{\includegraphics[width=\columnwidth]{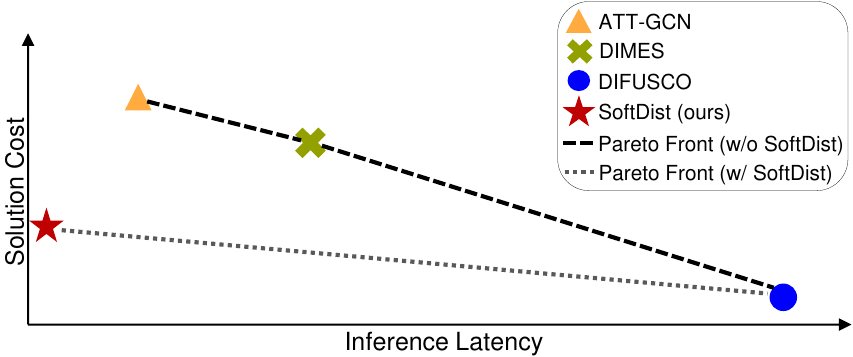}}
		\caption{Relative performance comparison of ML-based methods with and without SoftDist. \textit{Inference Latency} represents the heatmap generation time, with lower being better. \textit{Solution Quality} represents the effectiveness of TSP solutions generated through MCTS guided by the heatmaps, with higher being better.}
		\label{fig:pareto}
	\end{center}
	\vskip -0.3in
\end{figure}

\begin{figure*}[t]
	\vskip 0.2in
	\centering
	% Subfigure for Heatmap-Guided MCTS - Training Phase
	\begin{subfigure}[b]{0.45\textwidth}
		\includegraphics[width=\textwidth]{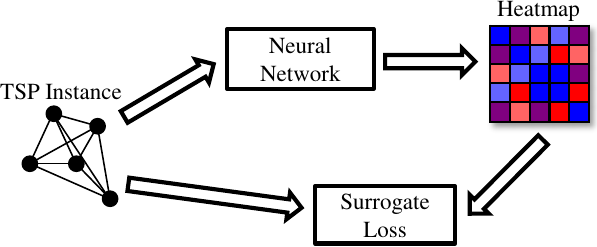}
		\caption{Training Phase}
		\label{fig:heatmap_train}
	\end{subfigure}
	\hfill % Space between figures
	% Subfigure for Heatmap-Guided MCTS - test phase
	\begin{subfigure}[b]{0.45\textwidth}
		\includegraphics[width=\textwidth]{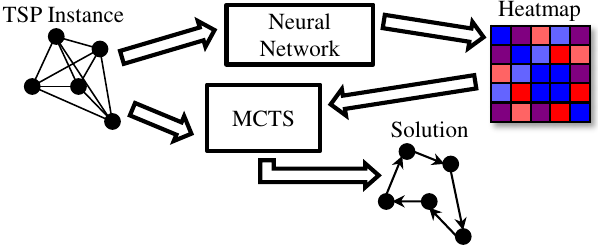}
		\caption{Test Phase}
		\label{fig:heatmap_test}
	\end{subfigure}
	\caption{Heatmap-Guided MCTS Phases.}
	\label{fig:heatmap_guided_mcts}
\end{figure*}

\begin{figure*}[t]
	\vskip 0.2in
	\centering
	% Subfigure for SoftDist-Guided MCTS - Training Phase
	\begin{subfigure}[b]{0.45\textwidth}
		\includegraphics[width=\textwidth]{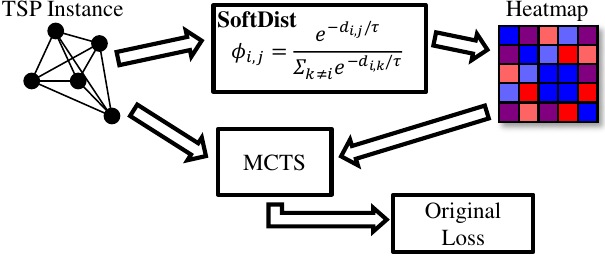}
		\caption{Training Phase}
		\label{fig:our_train}
	\end{subfigure}
	\hfill % Space between figures
	% Subfigure for SoftDist-Guided MCTS - test phase
	\begin{subfigure}[b]{0.45\textwidth}
		\includegraphics[width=\textwidth]{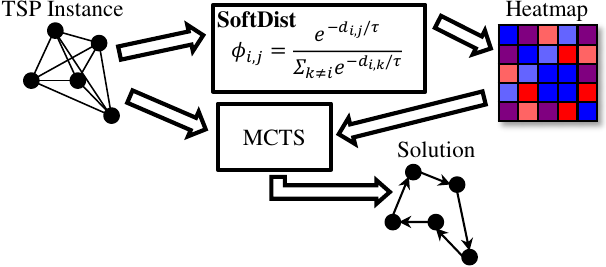}
		\caption{Test Phase}
		\label{fig:our_test}
	\end{subfigure}
	\caption{SoftDist-Guided MCTS Phases.}
	\label{fig:softdist_guided_mcts}
\end{figure*}

The traveling salesman problem (TSP) is a classic optimization challenge with significant applications in logistics, network design, and the broader field of operations research (OR). Traditionally addressed through exact algorithms like Concorde \cite{Helsgaun_2009} and heuristic algorithms such as LKH-3 \cite{Helsgaun_2017,Taillard_Helsgaun_2019}, recent years have seen a shift towards integrating machine learning (ML) for solving TSP, exemplified by \citet{bello,kool,costa}. However, these methods often lack scalability and become highly inefficient when applied to large-scale TSPs due to the exponentially growing action space, the quadratic computational complexity of self-attention mechanisms, and the issue of sparse rewards on large graphs \cite{transformer,bengio,rethink}.

Monte Carlo tree search (MCTS) is a versatile and adaptive algorithm widely applied across various domains \cite{Browne_mcts,Silver_mcts,Silver_mcts_2}. Its recent combination with ML, in efforts like ATT-GCN \cite{attgcn}, DIMES \cite{dimes}, UTSP \cite{utsp}, and DIFUSCO \cite{difusco}, represents a novel approach, known as heatmap-guided MCTS, for solving large-scale TSPs. These methods typically involve ML models generating heatmaps for TSP instances, assigning probabilities to each edge as potential parts of solutions, rather than directly generating TSP solutions. MCTS then utilizes these heatmaps as priors for edge selection to generate the final TSP solutions. However, the non-differentiable and time-consuming nature of the MCTS process complicates its direct integration into the training loss function of ML models. Consequently, there's a necessity for surrogate loss functions, typically designed heuristically, to facilitate ML model training. These surrogate losses aim to approximate the original TSP objective for a more feasible training process, but this heuristic approach can lead to inconsistencies between training (without MCTS integration) and testing phases (with MCTS), creating uncertainties in test scenario performances. This raises a critical question about the actual effectiveness of these ML-generated heatmaps in guiding MCTS. For a visual representation of the inconsistent training and test phases of ML-based heatmap generation methods, please refer to Figure \ref{fig:heatmap_guided_mcts}.

Moreover, considering the similarities between MCTS and LKH-3, such as their reliance on k-opt operations \cite{localoperator} and self-adaptive search strategies, and their implementation in C++ on CPU platforms, it's natural to conduct comparative experiments between heatmap-guided MCTS methods and LKH-3. LKH-3's status as a strong heuristic solver across various combinatorial problems prompts us to investigate: How effective is the heatmap-guided MCTS paradigm in comparison to LKH-3?

Aligning training and testing objectives is crucial. We introduce a straightforward baseline method, SoftDist, applying softmax to the TSP distance matrix. Its simplicity allows direct optimization using the original TSP objective via grid search, and importantly, it doesn't require hard-to-obtain supervision. For a visual representation of SoftDist's consistent training and test phases, refer to Figure \ref{fig:softdist_guided_mcts}. Our experiments demonstrate that SoftDist not only outperforms most complex ML-based heatmap generation methods in both solution quality and inference speed, but also achieves comparable performance to the fully-supervised DIFUSCO method \cite{difusco}, which requires hard-to-obtain supervision labels, as depicted in Figure \ref{fig:pareto}.

To facilitate a fair comparison between guided MCTS and LKH-3, we introduce the \textit{Score} metric, which evaluates MCTS performance relative to LKH-3 under identical hardware resources and time constraints. Our findings reveal that heatmap-guided MCTS methods significantly underperform compared to LKH-3, regardless of the allocated runtime or fine-tuning of MCTS parameters.

In this paper, we critically evaluate ML-guided heatmap generation and the heatmap-guided MCTS paradigm in large-scale TSPs, introducing key insights:
\begin{itemize}
	\item \textbf{Critical Evaluation:} We present the first comprehensive critique of both ML-based heatmap generation and the overall heatmap-guided MCTS paradigm for TSPs, highlighting critical insights into their effectiveness.
	\item \textbf{SoftDist Method:} We introduce SoftDist, an effective TSP heatmap generation method that surpasses complex ML methods, highlighting the ineffectiveness of current ML-based approaches and the necessity for theoretical validation in surrogate loss function designs.
	\item \textbf{\textit{Score} Metric:} We propose the \textit{Score} metric to evaluate the relative performance of heatmap-guided MCTS against the LKH-3 heuristic. This metric, utilized across different MCTS parameter settings and time budgets, reveals significant inefficiency of MCTS compared to LKH-3. It highlights the limited practical effectiveness of heatmap-guided MCTS in OR, despite the advancements in ML.
\end{itemize}

\section{Related Work}
This section reviews state-of-the-art methods for solving large-scale TSPs, all of which belong to the heatmap-guided MCTS paradigm. These methods are differentiated primarily by their heatmap generation approaches, categorized into supervised learning, unsupervised learning, and reinforcement learning.

\subsection{Supervised Learning}

ATT-GCN \cite{attgcn} employs a supervised model pre-trained on small-scale TSP instances (e.g., 20 or 50 nodes), based on a graph convolutional network \cite{joshi}. The model, once trained, generalizes to larger TSP instances through graph sampling techniques, creating sub-heatmaps that are merged to form a global heatmap for the original graph. This approach utilizes fixed-scale, small-scale TSP solutions as labels, enabling generalization to larger scales.

DIFUSCO \cite{difusco}, in contrast, adopts a fully-supervised training methodology, requiring TSP solutions for training instances at each scale. It introduces a graph-based diffusion model \cite{ho2020denoising,graikos2022diffusion}, treating TSP as a search for an optimal $\{0,1\}$-valued edge selection vector. Utilizing an anisotropic graph neural network \cite{bresson2018experimental}, the model iteratively denoises variables under supervision, with the final prediction serving as the heatmap.

\subsection{Unsupervised Learning}

UTSP \cite{utsp} employs an unsupervised approach, leveraging geometric scattering-based graph neural networks \cite{min2022can} to generate heatmaps. The method features a heuristically designed surrogate loss function optimized by the model. This unsupervised loss consists of two components: one that encourages the discovery of the shortest path, and another serving as a proxy for ensuring that the path forms a Hamiltonian Cycle covering all nodes.

\subsection{Reinforcement Learning}

DIMES \cite{dimes} adopts a reinforcement learning strategy, focusing on efficient sampling for REINFORCE-based gradient estimation \cite{williams1992simple}. Utilizing anisotropic graph neural networks \cite{bresson2018experimental}, it generates heatmaps, which are sampled using autoregressive factorization. This process creates a surrogate solution distribution approximating the TSP's true solution distribution, which is challenging to sample efficiently.

However, all of these methods have a significant limitation: the training of ML models for heatmap generation does not consider MCTS. Consequently, these models cannot ensure the quality of solutions that MCTS will derive from their heatmaps. 
This is concerning because the heatmap's effectiveness in guiding MCTS remains unpredictable, regardless of model performance during training. This disconnect between heatmap generation and its application in MCTS poses a critical challenge, emphasizing the need for alignment between these components.

\section{Preliminaries}

\subsection{Problem Definition}

Existing methods within heatmap-guided MCTS use the problem setting of 2D Euclidean TSP, and our study adheres to this established problem setting. We consider a TSP instance as an input graph with $n$ vertices in a two-dimensional space, represented by $s = \{x_{i}\}_{i=1}^{n}$ where $x_{i} \in [0,1]^{2}$. The goal is to find a permutation $\boldsymbol{\pi} = (\pi_{1}, \pi_{2}, ..., \pi_{n})$ that forms a tour, visiting each vertex once and returning to the start, with the objective to minimize the total path length $c(\boldsymbol{\pi})$, calculated as:
\begin{equation}
c(\boldsymbol{\pi}) = \left\| x_{\pi_n} - x_{\pi_1} \right\|_2 + \sum_{i=1}^{n-1} \left\| x_{\pi_i} - x_{\pi_{i+1}} \right\|_2 ,
\label{length}
\end{equation}
where  $\|\cdot\|_2$ denotes the $\ell_2$ norm.

\subsection{Heatmap Generation}

In the context of large-scale TSP, recent state-of-the-art approaches blend ML and OR, where ML models do not predict a solution (i.e., a permutation $\boldsymbol{\pi} = (\pi_{1}, \pi_{2}, ..., \pi_{n})$ of all the vertices) outright but alter the solution space distribution. Specifically, trained models predict an $n \times n$ heatmap $\Phi$, where $\Phi_{i,j}$ indicates the suitability of including edge $(i, j)$ in the solution. The optimization problem's objective is defined as:
\begin{equation}\label{original loss}
\mathcal{L}(\theta) = \mathbb{E}_{s \sim \mathcal{S}}\left[ \mathbb{E}_{\Phi \sim f_{\theta}(s)}\left[ \mathbb{E}_{\boldsymbol{\pi} \sim g(s,\Phi)} \left[ c\left( \boldsymbol{\pi} \right) \right] \right] \right],
\end{equation}
where $s$ represents an instance from distribution $\mathcal{S}$, $\theta$ is the trainable parameters of model $f$, $\boldsymbol{\pi}$ is the solution outputed by post-hoc search algorithm $g$ given $\Phi$, and $c(\boldsymbol{\pi} )$ is calculated based on Equation \ref{length}. 

Given the non-differentiable and computationally intensive nature of $\mathbb{E}_{\boldsymbol{\pi}  \sim g(s,\Phi)} \left[ c\left( \boldsymbol{\pi}  \right) \right]$, a surrogate loss $\ell\left(s,\Phi \right)$, which is both differentiable and easy to compute, is often employed, leading to a surrogate objective:
\begin{equation}\label{surrogate loss}
\mathcal{L}_{\textit{surrogate}}(\theta) = \mathbb{E}_{s \sim \mathcal{S}}\left[\mathbb{E}_{\Phi \sim f_{\theta}(s)}\left[ \ell\left(s,\Phi \right)  \right]  \right].
\end{equation}
This surrogate loss, designed heuristically, can take forms of supervised \cite{attgcn,difusco}, unsupervised \cite{utsp}, or reinforcement learning \cite{dimes}, where the optimized $\theta^{*}$ from minimizing $\mathcal{L}_{\textit{surrogate}}(\theta)$ is aimed to approximate the optimal $\theta$ obtained from the original loss, i.e., $\theta^{*} \approx \text{argmin}_{\theta}\mathbb{E}_{s \sim \mathcal{S}}\left[ \mathbb{E}_{\Phi \sim f_{\theta}(s)}\left[ \mathbb{E}_{\boldsymbol{\pi} \sim g(s,\Phi)} \left[ c\left( \boldsymbol{\pi} \right) \right] \right] \right]$. However, this approximation often lacks a rigorous theoretical foundation, making it uncertain whether minimizing the surrogate loss genuinely aligns with optimizing the original TSP objective. Consequently, despite optimizing $\theta^{*}$ for the surrogate loss, its efficacy in guiding MCTS to find optimal solutions during testing remains questionable. During inference, the output heatmap $\Phi^{*}$ from $f_{\theta^{*}}(s)$ is fed into the search algorithm $g$, yielding the solution $\boldsymbol{\pi}^{*} \sim g(s,\Phi^{*})$. This disconnect between training and test phases—where training focuses on heatmap generation without involving MCTS, while testing relies on MCTS guided by these heatmaps—highlights a potential misalignment in the approach, as depicted in Figure \ref{fig:heatmap_guided_mcts}.

\subsection{Monte Carlo Tree Search}

MCTS is utilized as a guided $k$-opt process, which iteratively refines a complete TSP solution $\boldsymbol{\pi}$ by alternating edge deletions and additions. The selection of edges during \(k\)-opt is influenced by a weight matrix \(W\) and an access matrix \(Q\), both of which are dynamically updated based on \(k\)-opt outcomes. Here, $W_{i, j}$ scores the suitability of edge $(i, j)$ in the solution, while $Q_{i, j}$ records the number of times edge $(i, j)$ is selected. Note that this section covers only the key aspects of MCTS. For a detailed understanding, please refer to \citet{target,attgcn,utsp}.

\paragraph{Initialization.}

The heatmap \(H\) initializes \(W\) (\(W_{i,j} = 100 \times H_{i,j}\)). The access matrix \(Q\) starts with all elements set to zero. Edge potential matrix \(Z\) guides the \(k\)-opt process, balancing exploitation and exploration. The edge potential $ Z_{i,j} $ is formulated as $Z_{i,j} = \frac{W_{i,j}}{\Omega_{i}} + \alpha \sqrt{\frac{\ln(M+1)}{Q_{i,j} + 1}}$, where $\Omega_{i}$, the average weight of edges connected to vertex $ i $, is $\Omega_{i} = \frac{\sum_{j \neq i} W_{i,j}}{\sum_{j \neq i} 1}$, $ \alpha $ balances exploitation and exploration, and $ M $ is the total number of actions sampled so far.

A random initial tour \(\boldsymbol{\pi}\) is constructed and optimized using 2-opt. The initial tour construction probability is formulated as $p(\boldsymbol{\pi}) = p(\pi_{1}) \prod_{i=2}^{n} p(\pi_{i} | \pi_{i-1})$, where $p(\pi_{i} | \pi_{i-1})$ is the conditional probability of choosing the next vertex, calculated by the edge potential:
\begin{equation}
p(\pi_{i} | \pi_{i-1}) = \frac{Z_{\pi_{i-1}, \pi_{i}}}{\sum_{l \in \mathbb{X}_{\pi_{i-1}}} Z_{\pi_{i-1}, l}},
\label{probability}
\end{equation}
with $ \mathbb{X}_{\pi_{i-1}} $ includes candidate vertices connected to $\pi_{i-1}$, selected based on their edge potential value.

\paragraph{$k$-opt Search.}

Each \(k\)-opt action is represented as a vertex decision sequence $(a_{1}, b_{1}, a_{2}, b_{2}, \ldots, a_{k}, b_{k}, a_{k+1})\) with \(a_{k+1} = a_{1}\). This sequence involves deleting \(k\) edges \((a_{i}, b_{i})\) and adding \(k\) new edges \((b_{i}, a_{i+1})\) for \(1 \leq i \leq k\). Given $b_{i}$, the subsequent vertex \(a_{i+1}\) is sampled based on Equation \ref{probability}. The tour \(\boldsymbol{\pi}\) is transformed into \(\boldsymbol{\pi}^\text{new}\), and metrics \(M\), \(Q_{b_i, a_{i+1}}\), and \(Q_{a_{i+1}, b_i}\) are updated.

\paragraph{Backpropagation.}

Upon obtaining a better solution $\boldsymbol{\pi}^{\text{new}}$ with $c(\boldsymbol{\pi}^{\text{new}}) < c(\boldsymbol{\pi})$, the weights of the newly added edges during the $k$-opt action are increased by $\beta\left[\text{exp}\left(\frac{c(\boldsymbol{\pi}) - c(\boldsymbol{\pi}^{\text{new}})}{c(\boldsymbol{\pi})}\right) - 1\right]$, where $\beta$ is the update rate.

\section{Proposed Baseline}

\subsection{Motivation}
Machine learning methods for generating TSP heatmaps usually rely on surrogate loss functions (Equation \ref{surrogate loss}) due to the computational challenges of the original loss (Equation \ref{original loss}). This substitution, often without theoretical justification, can lead to inconsistent performance in the test phase. This inconsistency is particularly concerning because MCTS, which is critical for determining the final solution, is not integrated during neural network training. 
In response, we introduced SoftDist, a baseline method that incorporates MCTS into the training process, thus directly optimizing the original TSP objective. However, the direct optimization of the original TSP loss presents challenges due to its non-differentiability and the time-consuming nature of deriving solutions via MCTS. Therefore, our aim with SoftDist is to simplify the optimization process by reducing the number of tunable parameters, effectively addressing these challenges.

\subsection{SoftDist Baseline}
We introduce a novel method for generating heatmaps, termed SoftDist, based on applying softmax to the distance matrix of a TSP instance. The heatmap \( \Phi \) allocates scores to each edge \( (i, j) \) as follows:
\begin{equation}
\Phi_{i,j} = \frac{e^{- d_{i,j}/\tau}}{\sum_{k \neq i} e^{- d_{i,k}/\tau}},
\label{softdist}
\end{equation}
where \( d_{i,j} = \left\| x_{i} - x_{j} \right\|_{2} \), and \( \tau \) is a parameter controlling the smoothness of the score distribution in \( \Phi \). This simplicity, with only one parameter to optimize, sets SoftDist apart from more complex models and aligns with our aim to simplify the optimization process. 

Moreover, our SoftDist method requires no supervision, significantly reducing its training complexity, especially beneficial for large-scale TSPs where obtaining high-quality labels is both expensive and challenging. Its inherent simplicity also ensures minimal hardware resource consumption, making it a highly practical option in various computational environments. This aspect of SoftDist underscores its efficiency and accessibility, further distinguishing it from more complex, resource-intensive ML models.

SoftDist's design prioritizes shorter edges while maintaining a balance between exploration and exploitation. This strategy is crucial for avoiding suboptimal, greedy solutions. By allocating scores inversely proportional to edge distances, moderated by $\tau$, SoftDist encourages structured exploration, aiding in superior solution discovery for large-scale TSPs.

During training, as illustrated in Figure \ref{fig:our_train}, SoftDist directly optimizes the original TSP objective, contrasting with the surrogate objectives used by other methods (see Figure \ref{fig:heatmap_train}). Owing to its single tunable parameter, $\tau$, SoftDist's optimization is straightforward, employing even the most basic optimization techniques like grid search. Once optimized, in the test phase, SoftDist generates heatmaps to guide MCTS, merging the training and test phases cohesively, as shown in Figure \ref{fig:our_test}. This alignment ensures that the heatmap's effectiveness in training directly translates to performance in testing.

\section{Proposed Metric}

\subsection{Motivation}

MCTS and LKH-3, both handcrafted heuristic algorithms, have several similarities, such as their reliance on local operators \cite{localoperator} and self-adaptive strategies for edge selection. This similarity lays the foundation for a comparative analysis between MCTS, particularly when guided by ML-generated heatmaps, and LKH-3, a leading heuristic solver for various routing problems. Past ML solvers for TSP  \cite{pointernetwork,bello,kool,costa,pomo,ma2021learning,ma2023neuopt} did not directly compare with LKH-3 due to differences in programming languages (e.g., Python vs. C++) and computational resources (e.g., GPU vs. CPU). Additionally, these ML solvers were designed as general-purpose solvers, typically independent of problem-specific features, while LKH-3 is a specialized solver relying on expert knowledge, tailored to specific types of problems, making direct comparisons unfair.
However, MCTS and LKH-3, both being problem-specific algorithms, share the same implementation environment, raising an important question: How does MCTS, even with external heatmap guidance, compare in performance to LKH-3?

\subsection{\textit{Score} Metric}\label{metriccalculate}

To establish an objective comparison between MCTS and LKH-3, we introduce the \textit{Score} metric, designed to assess the relative efficiency of MCTS compared to LKH-3 under identical programming and hardware conditions.
The \textit{Score} is calculated as the ratio of the performance gaps of LKH-3 and MCTS:
\begin{align}
\textit{Score} &= \frac{\textit{Gap}_{\text{LKH-3}}}{\textit{Gap}_{\text{MCTS}}},
\end{align}
where $\textit{Gap}_{\text{LKH-3}}=\frac{L_{\text{LKH-3}}}{L^{*}}-1$ and $\textit{Gap}_{\text{MCTS}}=\frac{L_{\text{MCTS}}}{L^{*}}-1$.
Among them, $L_{\text{LKH-3}}$ and $L_{\text{MCTS}}$ represent the solution lengths obtained by LKH-3 and MCTS, respectively. 
$L^{*}$ serves as the baseline for this comparison. Intuitively, \textit{Score} evaluates MCTS's relative efficiency, indicating the extent to which MCTS's performance is equivalent to that of LKH-3. A \textit{Score} above 100\% implies that MCTS is more efficient than LKH-3, while a score below 100\% indicates the opposite. This metric allows for an objective assessment of the performance of MCTS in relation to the efficacy of LKH-3.

\section{Experiments}
\begin{table*}[t]
	\caption{Results on large-scale TSP problems. Abbreviations: RL (Reinforcement learning), SL (Supervised learning), UL (Unsupervised learning), AS (Active search), G (Greedy decoding), S (Sampling decoding), and BS (Beam-search). $*$ indicates the baseline for performance gap calculation. $\dag$ indicates methods utilizing heatmaps provided by the original authors, with MCTS executed on our setup. $\#$ signifies methods without available heatmaps, requiring reproduction and potential overestimation of reported gaps due to issues found in their code. Some methods list two terms for \textit{Time}, corresponding to heatmap generation and MCTS runtimes, respectively. Baseline results (excluding those methods with MCTS) are sourced from \cite{attgcn,dimes}.}
	\vskip -0.15in
	\label{baseline}
	\begin{center}
		\begin{small}
			\begin{sc}
				\begin{adjustbox}{width=\textwidth}
					\begin{tabular}{lc|ccc|ccc|ccc}
						\toprule
						\multirow{2}{*}{Method} & \multirow{2}{*}{Type} & \multicolumn{3}{c|}{TSP-500} & \multicolumn{3}{c|}{TSP-1000} & \multicolumn{3}{c}{TSP-10000} \\
						& & Length $\downarrow$ & Gap $\downarrow$ & Time $\downarrow$ & Length $\downarrow$ & Gap $\downarrow$ & Time $\downarrow$ & Length $\downarrow$ & Gap $\downarrow$ & Time $\downarrow$ \\
						\midrule
						Concorde & OR(exact) & 16.55$^*$ & \textemdash & 37.66m & 23.12$^*$ & \textemdash & 6.65h & N/A & N/A & N/A \\
						Gurobi & OR(exact) & 16.55 & 0.00\% & 45.63h & N/A & N/A & N/A & N/A & N/A & N/A \\
						LKH-3 (default) & OR & 16.55 & 0.00\% & 46.28m & 23.12& 0.00\% & 2.57h & 71.78$^*$ & \textemdash & 8.8h \\
						LKH-3 (less trails) & OR & 16.55 & 0.00\% & 3.03m & 23.12 & 0.00\% & 7.73m & 71.79 & \textemdash & 51.27m \\
						Nearest Insertion & OR & 20.62 & 24.59\% & 0s & 28.96 & 25.26\% & 0s & 90.51 & 26.11\% & 6s \\
						Random Insertion & OR & 18.57 & 12.21\% & 0s & 26.12 & 12.98\% & 0s & 81.85 & 14.04\% & 4s \\
						Farthest Insertion & OR & 18.30 & 10.57\% & 0s & 25.72 & 11.25\% & 0s & 80.59 & 12.29\% & 6s \\
						\midrule
						EAN & RL+S & 28.63 & 73.03\% & 20.18m & 50.30 & 117.59\% & 37.07m & N/A & N/A & N/A \\
						EAN & RL+S+2-opt & 23.75 & 43.57\% & 57.76m & 47.73 & 106.46\% & 5.39h & N/A & N/A & N/A \\
						AM & RL+S & 22.64 & 36.84\% & 15.64m & 42.80 & 85.15\% & 63.97m & 431.58 & 501.27\% & 12.63m \\
						AM & RL+G & 20.02 & 20.99\% & 1.51m & 31.15 & 34.75\% & 3.18m & 141.68 & 97.39\% & 5.99m \\
						AM & RL+BS & 19.53 & 18.03\% & 21.99m & 29.90 & 29.23\% & 1.64h & 129.40 & 80.28\% & 1.81h \\
						GCN & SL+G & 29.72 & 79.61\% & 6.67m & 48.62 & 110.29\% & 28.52m & N/A & N/A & N/A \\
						GCN & SL+BS & 30.37 & 83.55\% & 38.02m & 51.26 & 121.73\% & 51.67m & N/A & N/A & N/A \\
						POMO+EAS-Emb & RL+AS & 19.24 & 16.25\% & 12.80h & N/A & N/A & N/A & N/A & N/A & N/A \\
						POMO+EAS-Lay & RL+AS & 19.35 & 16.92\% & 16.19h & N/A & N/A & N/A & N/A & N/A & N/A \\
						POMO+EAS-Tab & RL+AS & 24.54 & 48.22\% & 11.61h & 49.56 & 114.36\% & 63.45h & N/A & N/A & N/A \\
						\midrule
						DIFUSCO$^{\#}$ & SL+MCTS & 16.63 & 0.51\% & \begin{tabular}[c]{@{}c@{}}3.61m+\\1.67m\end{tabular} & 23.39 & 1.18\% & \begin{tabular}[c]{@{}c@{}}11.86m+\\3.34m\end{tabular} & 73.76 & 2.77\% & \begin{tabular}[c]{@{}c@{}}28.51m+\\16.87m\end{tabular} \\
						\midrule
						ATT-GCN$^{\dag}$ & SL+MCTS & 16.82 & 1.64\% & \begin{tabular}[c]{@{}c@{}}0.52m+\\1.67m\end{tabular} & 23.67 & 2.37\% & \begin{tabular}[c]{@{}c@{}}0.73m+\\3.34m\end{tabular} & 74.50 & 3.80\% & \begin{tabular}[c]{@{}c@{}}4.16m+\\16.77m\end{tabular} \\
						DIMES$^{\dag}$ & RL+MCTS & 16.84 & 1.77\% & \begin{tabular}[c]{@{}c@{}}0.97m+\\1.67m\end{tabular} & 23.68 & 2.44\% & \begin{tabular}[c]{@{}c@{}}2.08m+\\3.34m\end{tabular} & 74.10 & 3.23\% & \begin{tabular}[c]{@{}c@{}}4.65m+\\16.77m\end{tabular} \\
						UTSP$^{\dag}$ & UL+MCTS & 17.11 & 3.41\% & \begin{tabular}[c]{@{}c@{}}1.37m+\\1.67m\end{tabular} & 24.14 & 4.40\% & \begin{tabular}[c]{@{}c@{}}3.35m+\\3.34m\end{tabular} & \textemdash & \textemdash & \textemdash \\
						Ours & SoftDist+MCTS & \textbf{16.78} & \textbf{1.44\%} & \begin{tabular}[c]{@{}c@{}}\textbf{0.00m+}\\\textbf{1.67m}\end{tabular} & \textbf{23.63} & \textbf{2.20\%} & \begin{tabular}[c]{@{}c@{}}\textbf{0.00m}+\\\textbf{3.34m}\end{tabular} & \textbf{74.03} & \textbf{3.13\%} & \begin{tabular}[c]{@{}c@{}}\textbf{0.00m+}\\\textbf{16.78m}\end{tabular} \\
						\bottomrule
					\end{tabular}
				\end{adjustbox}
			\end{sc}
		\end{small}
	\end{center}
	\vskip -0.08in
\end{table*}

\begin{table*}[t]
	\caption{Resource consumption and \textit{Score} comparison of various methods on TSPs. \textit{Score} measures the efficiency relative to LKH-3. Detailed metric calculations are in Section \ref{metriccalculate}. The definitions of notations $\dag$ and $\#$ are explained in Table \ref{baseline}.}
	\label{metric}
	\vskip 0.15in
	\begin{center}
		\begin{small}
			\begin{sc}
				\begin{tabular}{lcc|ccc}
					\toprule
					\multirow{2}{*}{Method} & \multirow{2}{*}{Supervision} & \multirow{2}{*}{Hardware} & \multicolumn{3}{c}{Score $\uparrow$} \\
					& & & TSP-500 & TSP-1000 & TSP-10000 \\
					\midrule
					ATT-GCN$^{\dag}$ & \ding{51} & GTX 1080 Ti GPU & 0.74\% & 3.87\% & 24.66\% \\
					DIMES$^{\dag}$ & \ding{55} & NVIDIA P100 GPU & 0.68\% & 3.75\% & 28.99\% \\
					UTSP$^{\dag}$ & \ding{55} & NVIDIA V100 GPU & 0.35\% & 2.08\% & \textemdash \\
					DIFUSCO$^{\#}$ & \ding{51} & 8$\times$NVIDIA V100 GPUs & 2.39\% & 7.78\% & 33.82\% \\
					SoftDist & \ding{55} & NVIDIA A100 GPU & 0.84\% & 4.17\% & 29.88\% \\
					\bottomrule
				\end{tabular}
			\end{sc}
		\end{small}
	\end{center}
	\vskip -0.1in
\end{table*}

\begin{table*}[t]
	\caption{Comparison of MCTS performance under the different settings by \cite{utsp} for TSP-500 and TSP-1000. The \textit{Score} metric is detailed in Section \ref{metriccalculate}. Definitions of superscript notations $*$, $\dag$, and $\#$ are provided in Table \ref{baseline}.}
	\label{different param}
	\vskip 0.15in
	\begin{center}
		\begin{small}
			\begin{sc}
				\begin{adjustbox}{width=\textwidth}
					\begin{tabular}{l|cccc|cccc}
						\toprule
						\multirow{2}{*}{Method} & \multicolumn{4}{c|}{TSP-500} & \multicolumn{4}{c}{TSP-1000} \\
						& Length $\downarrow$ & Gap $\downarrow$ & Time $\downarrow$ & Score $\uparrow$ & Length $\downarrow$ & Gap $\downarrow$ & Time $\downarrow$ & Score $\uparrow$ \\
						\midrule
						LKH-3 (default) & 16.55$^*$ & 0.00\% & 46.28m & \textemdash & 23.12$^*$ & 0.00\% & 2.57h & \textemdash \\
						\midrule
						ATT-GCN$^{\dag}$ & 16.72 & 1.02\% & 0.52m+0.67m & 5.38\% & 23.48 & 1.58\% & 0.73m+1.43m & 13.77\% \\
						DIMES$^{\dag}$ & 16.75 & 1.26\% & 0.97m+0.68m & 4.35\% & 23.61 & 2.11\% & 2.08m+1.45m & 10.29\%\\
						DIFUSCO$^{\#}$ & 16.69 & 0.90\% & 3.61m+0.68m & 6.12\% & 23.48 & 1.56\% & 11.86m+1.43m & 13.93\%\\
						UTSP$^{\dag}$ & 16.73 & 1.09\% & 1.37m+0.68m & 5.05\% & 23.50 & 1.65\% & 3.35m+1.45m & 13.18\%\\
						SoftDist & 16.72 & 1.03\% & 0.00m+0.68m & 5.32\% & 23.52 & 1.73\% & 0.00m+1.44m & 12.56\%\\
						Zeros & 16.72 & 1.06\% & 0.00m+0.68m & 5.20\% & 23.55 & 1.85\% & 0.00m+1.44m & 11.72\% \\
						\bottomrule
					\end{tabular}
				\end{adjustbox}
			\end{sc}
		\end{small}
	\end{center}
	\vskip -0.1in
\end{table*}

\begin{figure*}[t]
	\vskip 0.2in
	\centering
	\begin{subfigure}[b]{0.33\textwidth}
		\includegraphics[width=\textwidth]{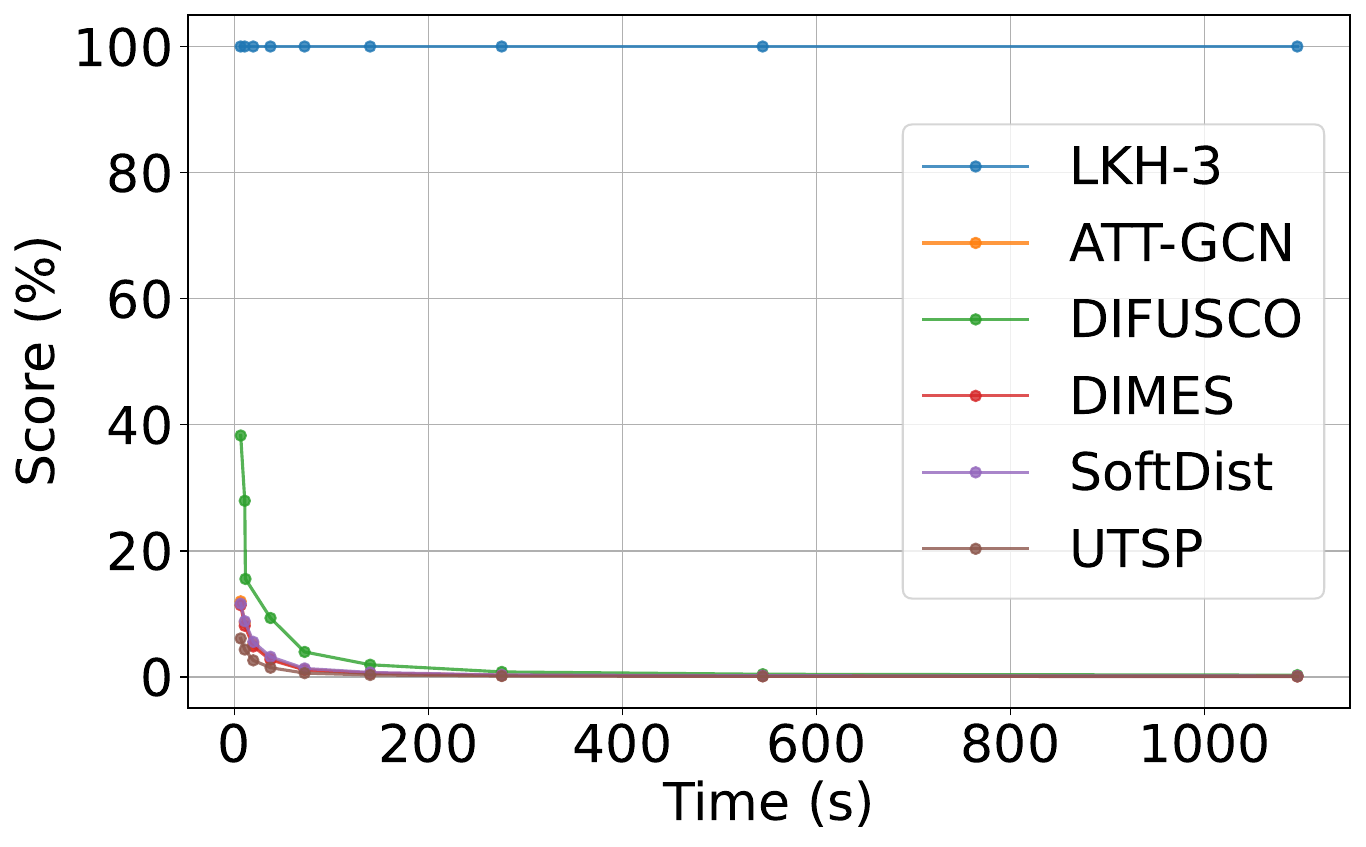}
		\caption{TSP-500 with default MCTS settings.}
		\label{fig:tsp500_default}
	\end{subfigure}
	\hfill
	\begin{subfigure}[b]{0.33\textwidth}
		\includegraphics[width=\textwidth]{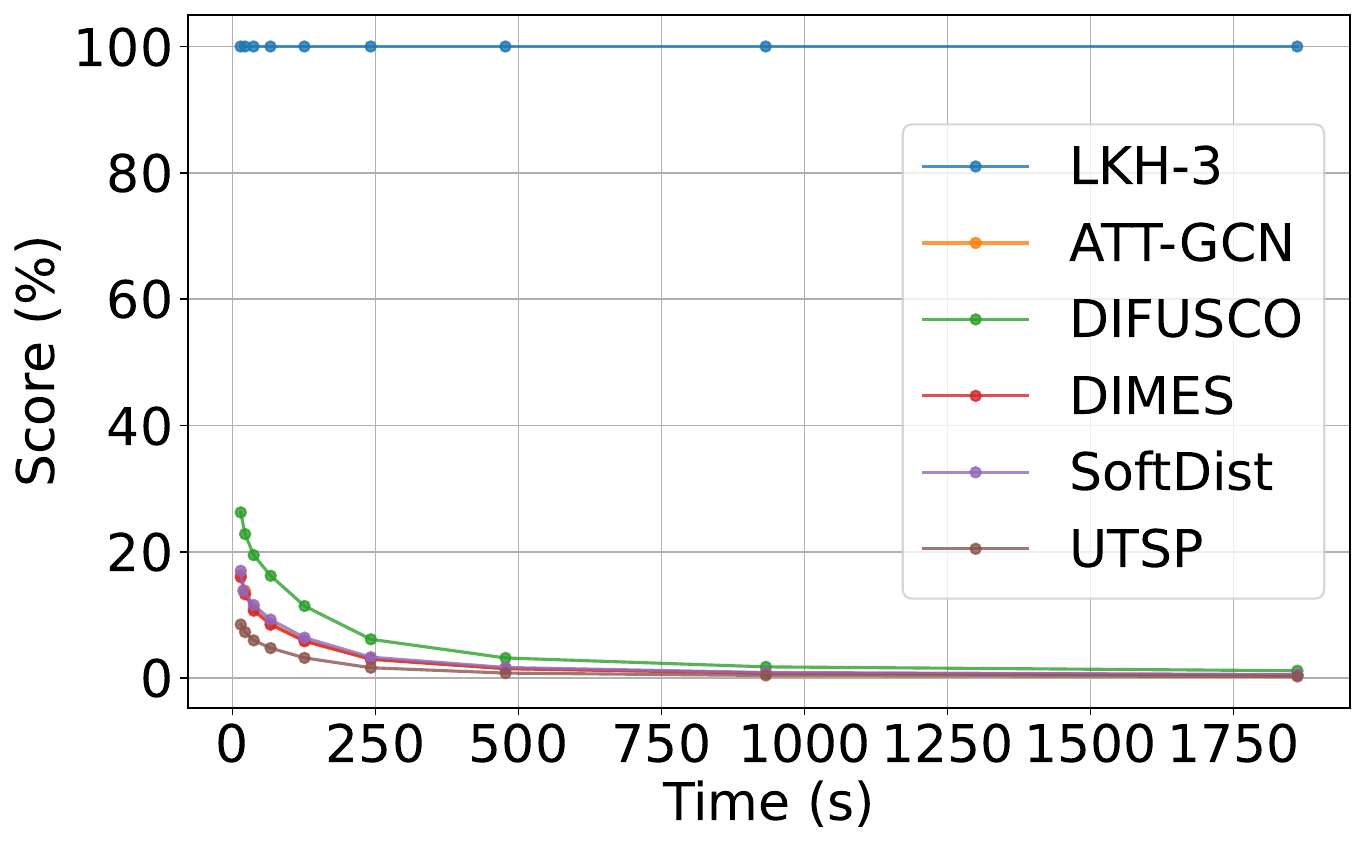}
		\caption{TSP-1000 with default MCTS settings.}
		\label{fig:tsp1000_default}
	\end{subfigure}
	\hfill
	\begin{subfigure}[b]{0.33\textwidth}
		\includegraphics[width=\textwidth]{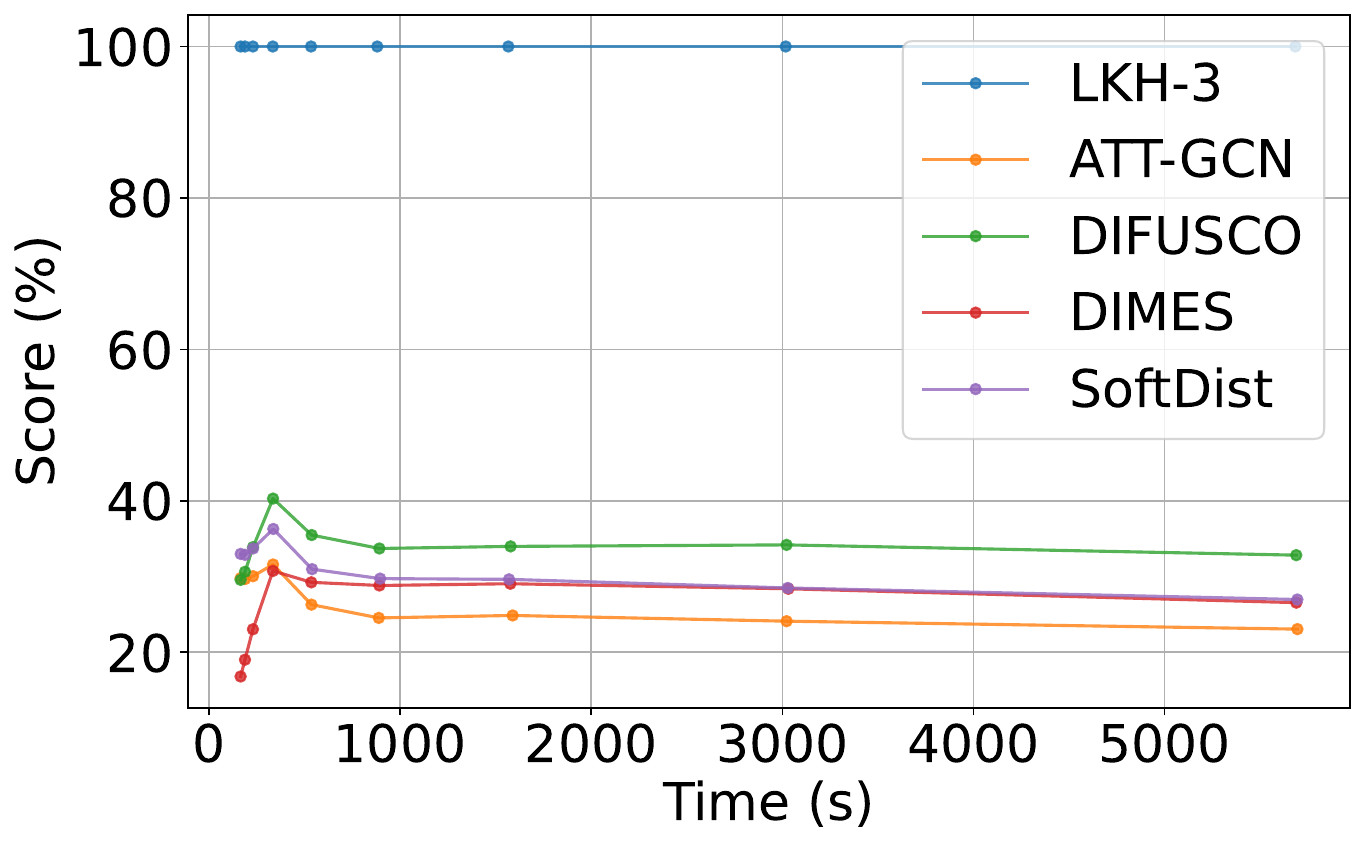}
		\caption{TSP-10000 with default MCTS settings.}
		\label{fig:tsp10000_default}
	\end{subfigure}
	\caption{Performance of MCTS under default settings for TSP-500, TSP-1000, and TSP-10000.}
	\label{fig:tsp_default}
\end{figure*}

\subsection{Experimental Settings}

\paragraph{Data sets} We follow the data generation as seen in \citet{kool}, creating TSP problems named TSP-500/1000/10000, where TSP-$n$ represents instances with $n$ nodes. We generate 1024 two-dimensional Euclidean TSP instances for TSP-500/1000 and 128 instances for TSP-10000 for parameter searching, using a random seed of 1234. 
% Nodes for each instance are uniformly distributed over a unit square. 
For testing purposes, we utilize test instances generated by \citet{attgcn}, following the approach used in \citet{dimes,utsp,difusco}. 
% The test set includes 128 instances for TSP-500/1000 and 16 instances for TSP-10000.

\paragraph{SoftDist temperature} For determining the optimal SoftDist temperature parameter $\tau$ defined in Equation \ref{softdist}, we conduct a grid search on the generated training instances. We identify the optimal temperatures for heatmap generation to be 0.0066 for TSP-500, 0.0051 for TSP-1000, and 0.0018 for TSP-10000. Detailed results of the grid search are presented in Appendix \ref{grid_search_appendix}.

\paragraph{MCTS parameters} We maintain default settings for all MCTS parameters, including $\alpha$, $\beta$, time budgets, etc., consistent with approaches in \citet{attgcn,dimes,difusco}. 
These default parameters are fixed across various problem scales. Notably, \citet{utsp} uses a different approach, applying different parameter settings for different scale TSPs. For a fair comparison, we aligned the parameter settings in \citet{utsp} with those used in the other referenced works.

\paragraph{Evaluation metrics} In our model comparison, we report the average tour length (\textit{Length}), average performance gap (\textit{Gap}), and average inference latency time (\textit{Time}) in Table \ref{baseline}. \textit{Length} (lower is better) represents the average length of the predicted tour for each graph in the test set. \textit{Gap} (smaller is better) measures the average relative performance gap in solution length compared to a baseline method. \textit{Time} (shorter is better) denotes the total clock time to generate solutions for all test instances, reported in seconds (s), minutes (m), or hours (h). For our proposed metric \textit{Score}, we follow the default setting of LKH-3 in \citet{kool,dimes} by setting the maximum of trials to 10000 and align the search time with MCTS by adjusting the number of runs.

\paragraph{Hardware} Our SoftDist heatmap generation is performed on an NVIDIA A100 GPU. Due to its simplicity, the inference time is negligible (e.g., $< 0.1$ seconds), and the GPU type has minimal impact. For fairness in comparison, all MCTS computations are conducted on an AMD EPYC 7V13 64-Core CPU @ 2.45GHz. We use 64 threads for TSP-500 and TSP-1000, and 16 threads for TSP-10000, following the setup in \citet{dimes}. To evaluate all methods under the \textit{Score} metric, LKH-3 is also run on the same CPU with the same thread count.

\subsection{Results and Analyses} \label{results and analyses}

\paragraph{\textit{How effective are the heatmaps generated by deep neural networks?}} In Table \ref{baseline}, we compare our SoftDist approach with state-of-the-art methods on TSP-500, TSP-1000, and TSP-10000. Notably, our simple baseline significantly outperforms most existing neural solvers across all three problem sizes, achieving lower gaps with negligible heatmap generation time under the same MCTS settings. Specifically, SoftDist demonstrates a performance gap of just \textbf{1.44\%} for TSP-500, \textbf{2.20\%} for TSP-1000, and \textbf{3.13\%} for TSP-10000, highlighting its effectiveness in heatmap generation. Moreover, SoftDist's simplicity and hardware efficiency make it less resource-intensive than other methods. An exception is DIFUSCO \cite{difusco}, which achieves Pareto optimality alongside SoftDist, showing lower performance gaps but at the expense of considerably longer heatmap generation times. Specifically, for TSP-500, DIFUSCO takes 3.61 minutes for heatmap generation, which is 2.2 times the MCTS search time. For TSP-1000, it requires 11.86 minutes, approximately 3.6 times the MCTS search duration. In the case of TSP-10000, the heatmap generation time is 28.51 minutes, about 1.7 times the MCTS search time. By contrast, our SoftDist method generates heatmaps in less than \textbf{0.1 seconds} for TSP-10000 and even under \textbf{0.01 seconds} for TSP-500 and TSP-1000. For a visual representation of the dominance and Pareto relationship between these methods exemplified in TSP-10000, please refer to Figure \ref{fig:pareto}. Moreover, DIFUSCO requires high-quality labels for each TSP scale and is trained directly on large-scale TSPs, consuming substantial hardware resources, as shown in Table \ref{metric}. Unlike DIFUSCO, SoftDist does not need ground truth solutions and is hardware-friendly. 

The aim of introducing SoftDist is not necessarily to outperform other models but to provide a benchmark for evaluating the effectiveness of ML-based approaches in TSP heatmap generation. Our findings, particularly the notable performance of SoftDist, illuminate a significant shortcoming in current ML methods: their dependence on surrogate loss functions. These heuristic loss functions, while simplifying training, often lack theoretical grounding, leading to a gap in performance during test phases, especially when integrated with MCTS. Such a discrepancy validates our position that ML-based heatmap generation in large-scale TSPs may not be as effective as anticipated. This necessitates a more aligned approach in heatmap generation, harmonizing training and test phases, and fostering the development of more consistent and theoretically robust ML models for TSPs.

\paragraph{\textit{How effective is the heatmap-guided MCTS paradigm for large-scale TSPs?}} Our experimental analysis, employing the \textit{Score} metric in Table \ref{metric}, critically evaluates the practical effectiveness of the heatmap-guided MCTS paradigm in large-scale TSPs. The results clearly highlight the paradigm's limitations, particularly when compared to LKH-3 under the same computational resources and time constraints. 
Across various TSP scales, MCTS consistently achieves \textit{Scores} that are significantly low, indicating its substantial underperformance.
For example, MCTS's performance is less than one-tenth for TSP-500 and TSP-1000, and about a third for TSP-10000, relative to LKH-3. These findings emphasize a significant performance gap, reinforcing our position that despite advancements in TSP through deep learning, traditional heuristic methods like LKH-3 still maintain a significant advantage in efficiency and applicability for these problems. The pronounced underperformance of heatmap-guided MCTS, despite significant resource investment in training and inference, motivates our call for future research to narrow this gap and explore possibilities to potentially outperform established heuristics like LKH-3.

\paragraph{\textit{How does MCTS perform under varying parameter settings?}} We evaluate different MCTS settings proposed by \citet{utsp}, which vary based on TSP scale and incorporate randomness into the search process, noted for improved performance. For the detailed implementation of these settings, please refer to \citet{utsp}. Table \ref{different param} indicates that, except for DIFUSCO, all methods show performance improvements even with reduced search time budgets, compared with the default MCTS settings in Table \ref{baseline}. This improvement, consistent across methods even including a zero-input heatmap\footnote{Actually, all elements of the heatmap are set to $10^{-10}$ to avoid division by zero errors.} baseline, which is referred to as Zeros in Table \ref{different param}, indicates the reduced influence of the heatmap and the enhanced impact of fine-tuned MCTS parameters on search efficiency. However, despite the parameter optimization, MCTS methods do not outperform LKH-3 in effectiveness, which means that for large-scale TSP problems, LKH-3 remains a more preferable choice. The \textit{Score} of MCTS methods, even lower than 7\% for TSP-500 and 14\% for TSP-1000, further highlights the superior efficiency of LKH-3 over MCTS methods. This aligns with our position that heatmap-guided MCTS, despite parameter optimization, remains less practical and effective compared to LKH-3 for large-scale TSP problems.

\paragraph{\textit{How does MCTS perform under varying time budgets?}} 
In Figure \ref{fig:tsp_default}, we examine the performance of heatmap-guided MCTS under various time budgets, and the results show that MCTS methods consistently underperform compared to LKH-3. Specifically, as time budgets increase for TSP-500 and TSP-1000, the \textit{Score} for MCTS methods approaches zero, indicating a persistent and significant performance gap for MCTS even as LKH-3 nearly optimizes the solution. 
For the experimental results of MCTS under \citet{utsp}'s settings, which show similar trends to the default settings, please refer to Appendix \ref{appendix:utsp_results}.

For TSP-10000, we observe that longer time budgets do not exhibit rapid convergence to zero, as seen with TSP-500 and TSP-1000. Instead, the score enters a plateau phase, showing only gradual changes. This pattern suggests that both MCTS and LKH-3 do not significantly enhance solution quality with increased time, indicating a slower optimization process for larger-scale problems. Notably, the performance curves of heatmap-guided MCTS methods exhibit a turning point, and within a time budget of around 230 seconds or less, our SoftDist method demonstrates the most effective performance among the heatmap-guided MCTS approaches. Nevertheless, all methods, including SoftDist, still fall significantly short of LKH-3's performance. In summary, these findings affirm our position that the heatmap-guided MCTS paradigm, despite its innovative approach, shows limited practical effectiveness compared to LKH-3 across various scenarios, whether with ample or limited time budgets.

\section{Conclusion, Discussion and Future Work}

\paragraph{Conclusion.} This paper presents, for the first time, a critical evaluation of ML-based heatmap generation and the heatmap-guided MCTS paradigm in large-scale TSPs, aligning with our position on their limitations. We introduced SoftDist, a simple yet effective baseline, outperforming more complex ML-based heatmap generation methods in solution quality and inference speed. SoftDist aims not to surpass but to provoke a reconsideration of current ML-based methods. Additionally, we proposed a novel metric \textit{Score} for evaluating the relative effectiveness of the guided MCTS compared to LKH-3, revealing a significant performance gap, underscoring the limited practical effectiveness of the heatmap-guided MCTS approach. We believe our proposed baseline and metric can serve as valuable benchmarks for future research in this domain.

\paragraph{Discussion.} A key issue with heatmap-guided MCTS is its reliance on surrogate loss functions, which do not directly optimize the original TSP loss and lack a rigorous theoretical foundation, resulting in uncertain performance during test phases. Moreover, the reliance on post-hoc search methods like MCTS contradicts the original goal of using ML in OR, which is to develop generalizable, autonomous, problem-agnostic algorithms. This continued dependence on handcrafted, problem-specific search strategies is contrary to the intended automation and generalizability of ML solutions in OR.

\paragraph{Future work.} Future research should focus on developing more effective heatmap generation methods with a theoretical basis for their loss functions. Additionally, exploring end-to-end solution generation methods, which generate solutions directly without complex postprocessing steps, despite their current performance lagging behind MCTS-based methods, offers another promising direction.
% Though currently less effective than MCTS-based methods, these non-MCTS approaches could pave the way towards more autonomous ML-driven solutions in operational research.
% \newpage
\section*{Impact Statement}
Our research critically evaluates ML-guided heatmap generation and the heatmap-guided MCTS paradigm in large-scale TSPs. The potential broader impact of this work lies in advancing the field of ML and OR, especially in complex problem-solving like logistics and network design. We highlight the need for more theoretically robust approaches in ML and the exploration of efficient, autonomous ML methods for combinatorial problems. These advancements could lead to more sustainable and effective solutions in various industries, while also underscoring the importance of aligning theoretical soundness with practical applicability in ML research and applications.
% In the unusual situation where you want a paper to appear in the
% references without citing it in the main text, use \nocite
\nocite{langley00}
\bibliography{example_paper}

\begin{thebibliography}{44}
\providecommand{\natexlab}[1]{#1}
\providecommand{\url}[1]{\texttt{#1}}
\expandafter\ifx\csname urlstyle\endcsname\relax
  \providecommand{\doi}[1]{doi: #1}\else
  \providecommand{\doi}{doi: \begingroup \urlstyle{rm}\Url}\fi

\bibitem[Applegate et~al.(2006)Applegate, Bixby, Chvatál, and
  Cook]{Applegat2007}
Applegate, D.~L., Bixby, R.~E., Chvatál, V., and Cook, W.~J.
\newblock \emph{The Traveling Salesman Problem: A Computational Study}.
\newblock Princeton University Press, 2006.

\bibitem[Applegate et~al.(2009)Applegate, Bixby, Chvátal, Cook, Espinoza,
  Goycoolea, and Helsgaun]{Helsgaun_2009}
Applegate, D.~L., Bixby, R.~E., Chvátal, V., Cook, W., Espinoza, D.~G.,
  Goycoolea, M., and Helsgaun, K.
\newblock Certification of an optimal {TSP} tour through 85,900 cities.
\newblock \emph{Operations Research Letters}, 37\penalty0 (1):\penalty0
  11–--15, 2009.

\bibitem[Bello et~al.(2016)Bello, Pham, Le, Norouzi, and Bengio]{bello}
Bello, I., Pham, H., Le, Q.~V., Norouzi, M., and Bengio, S.
\newblock Neural combinatorial optimization with reinforcement learning.
\newblock \emph{arXiv preprint arXiv:1611.09940}, 2016.

\bibitem[Bengio et~al.(2021)Bengio, Lodi, and Prouvost]{bengio}
Bengio, Y., Lodi, A., and Prouvost, A.
\newblock Machine learning for combinatorial optimization: A methodological
  tour d’horizon.
\newblock \emph{European Journal of Operational Research}, 290\penalty0
  (2):\penalty0 405--421, 2021.

\bibitem[Bresson \& Laurent(2018)Bresson and Laurent]{bresson2018experimental}
Bresson, X. and Laurent, T.
\newblock An experimental study of neural networks for variable graphs.
\newblock In \emph{ICLR Workshop}, 2018.

\bibitem[Browne et~al.(2012)Browne, Powley, Whitehouse, Lucas, Cowling,
  Rohlfshagen, Tavener, Perez, Samothrakis, and Colton]{Browne_mcts}
Browne, C.~B., Powley, E., Whitehouse, D., Lucas, S.~M., Cowling, P.~I.,
  Rohlfshagen, P., Tavener, S., Perez, D., Samothrakis, S., and Colton, S.
\newblock A survey of {Monte Carlo} tree search methods.
\newblock \emph{IEEE Transactions on Computational Intelligence and AI in
  Games}, 4\penalty0 (1):\penalty0 1--43, 2012.

\bibitem[Chen \& Tian(2019)Chen and Tian]{chen_tian}
Chen, X. and Tian, Y.
\newblock Learning to perform local rewriting for combinatorial optimization.
\newblock In \emph{NeurIPS}, pp.\  6278--6289, 2019.

\bibitem[Coulom(2006)]{Coulom_mcts}
Coulom, R.
\newblock Efficient selectivity and backup operators in {Monte-Carlo} tree
  search.
\newblock In \emph{ICCG}, pp.\  72--83, 2006.

\bibitem[da~Costa et~al.(2020)da~Costa, Rhuggenaath, Zhang, and Akçay]{costa}
da~Costa, P., Rhuggenaath, J., Zhang, Y., and Akçay, A.~E.
\newblock Learning 2-opt heuristics for the traveling salesman problem via deep
  reinforcement learning.
\newblock In \emph{ACML}, pp.\  465--480, 2020.

\bibitem[Drori et~al.(2020)Drori, Kharkar, Sickinger, Kates, Ma, Ge, Dolev,
  Dietrich, Williamson, and Udell]{realworld}
Drori, I., Kharkar, A., Sickinger, W.~R., Kates, B., Ma, Q., Ge, S., Dolev, E.,
  Dietrich, B.~L., Williamson, D.~P., and Udell, M.
\newblock Learning to solve combinatorial optimization problems on real-world
  graphs in linear time.
\newblock \emph{IEEE International Conference on Machine Learning and
  Applications}, pp.\  19--24, 2020.

\bibitem[Fu et~al.(2019)Fu, Qiu, Qiu, and Zha]{target}
Fu, Z.-H., Qiu, K.-B., Qiu, M., and Zha, H.
\newblock Targeted sampling of enlarged neighborhood via {Monte Carlo} tree
  search for {TSP}.
\newblock 2019.

\bibitem[Fu et~al.(2021)Fu, Qiu, and Zha]{attgcn}
Fu, Z.-H., Qiu, K.-B., and Zha, H.
\newblock Generalize a small pre-trained model to arbitrarily large {TSP}
  instances.
\newblock In \emph{AAAI}, pp.\  7474--7482, 2021.

\bibitem[Graikos et~al.(2022)Graikos, Malkin, Jojic, and
  Samaras]{graikos2022diffusion}
Graikos, A., Malkin, N., Jojic, N., and Samaras, D.
\newblock Diffusion models as plug-and-play priors.
\newblock In \emph{NeurIPS}, pp.\  14715--14728, 2022.

\bibitem[Helsgaun(2017)]{Helsgaun_2017}
Helsgaun, K.
\newblock \emph{An Extension of the Lin-Kernighan-Helsgaun TSP Solver for
  Constrained Traveling Salesman and Vehicle Routing Problems: Technical
  report}.
\newblock Roskilde Universitet, 2017.

\bibitem[Ho et~al.(2020)Ho, Jain, and Abbeel]{ho2020denoising}
Ho, J., Jain, A., and Abbeel, P.
\newblock Denoising diffusion probabilistic models.
\newblock In \emph{NeurIPS}, pp.\  6840--6851, 2020.

\bibitem[Hottung \& Tierney(2022)Hottung and Tierney]{hottung}
Hottung, A. and Tierney, K.
\newblock Neural large neighborhood search for routing problems.
\newblock \emph{Artificial Intelligence}, 313:\penalty0 103786, 2022.

\bibitem[Joshi et~al.(2019)Joshi, Laurent, and Bresson]{joshi}
Joshi, C.~K., Laurent, T., and Bresson, X.
\newblock An efficient graph convolutional network technique for the travelling
  salesman problem.
\newblock \emph{ArXiv}, abs/1906.01227, 2019.

\bibitem[Joshi et~al.(2022)Joshi, Cappart, Rousseau, and Laurent]{rethink}
Joshi, C.~K., Cappart, Q., Rousseau, L.-M., and Laurent, T.
\newblock Learning the travelling salesperson problem requires rethinking
  generalization.
\newblock \emph{Constraints}, 27\penalty0 (1-2):\penalty0 70--98, 2022.

\bibitem[Kim et~al.(2022)Kim, Park, and Park]{sym-nco}
Kim, M., Park, J., and Park, J.
\newblock {Sym-NCO}: Leveraging symmetricity for neural combinatorial
  optimization.
\newblock In \emph{NeurIPS}, pp.\  1936--1949, 2022.

\bibitem[Kool et~al.(2019)Kool, van Hoof, and Welling]{kool}
Kool, W., van Hoof, H., and Welling, M.
\newblock Attention, learn to solve routing problems!
\newblock In \emph{ICLR}, pp.\  1--25, 2019.

\bibitem[Kool et~al.(2022)Kool, van Hoof, Gromicho, and Welling]{dpdp}
Kool, W., van Hoof, H., Gromicho, J., and Welling, M.
\newblock Deep policy dynamic programming for vehicle routing problems.
\newblock In \emph{CPAIOR}, pp.\  190--213, 2022.

\bibitem[Kwon et~al.(2020)Kwon, Choo, Kim, Yoon, Gwon, and Min]{pomo}
Kwon, Y.-D., Choo, J., Kim, B., Yoon, I., Gwon, Y., and Min, S.
\newblock {POMO}: Policy optimization with multiple optima for reinforcement
  learning.
\newblock In \emph{NeurIPS}, pp.\  21188--21198, 2020.

\bibitem[Lu et~al.(2020)Lu, Zhang, and Yang]{lu}
Lu, H., Zhang, X., and Yang, S.
\newblock A learning-based iterative method for solving vehicle routing
  problems.
\newblock In \emph{ICLR}, pp.\  1--15, 2020.

\bibitem[Ma et~al.(2021{\natexlab{a}})Ma, Li, Cao, Song, Zhang, Chen, and
  Tang]{dual}
Ma, Y., Li, J., Cao, Z., Song, W., Zhang, L., Chen, Z., and Tang, J.
\newblock Learning to iteratively solve routing problems with dual-aspect
  collaborative transformer.
\newblock In \emph{NeurIPS}, pp.\  11096--11107, 2021{\natexlab{a}}.

\bibitem[Ma et~al.(2021{\natexlab{b}})Ma, Li, Cao, Song, Zhang, Chen, and
  Tang]{ma2021learning}
Ma, Y., Li, J., Cao, Z., Song, W., Zhang, L., Chen, Z., and Tang, J.
\newblock Learning to iteratively solve routing problems with dual-aspect
  collaborative transformer.
\newblock In \emph{NeurIPS}, pp.\  11096--11107, 2021{\natexlab{b}}.

\bibitem[Ma et~al.(2023)Ma, Cao, and Chee]{ma2023neuopt}
Ma, Y., Cao, Z., and Chee, Y.~M.
\newblock Learning to search feasible and infeasible regions of routing
  problems with flexible neural k-opt.
\newblock In \emph{NeurIPS}, 2023.

\bibitem[Min et~al.(2022)Min, Wenkel, Perlmutter, and Wolf]{min2022can}
Min, Y., Wenkel, F., Perlmutter, M., and Wolf, G.
\newblock Can hybrid geometric scattering networks help solve the maximum
  clique problem?
\newblock In \emph{NeurIPS}, pp.\  22713--22724, 2022.

\bibitem[Min et~al.(2023)Min, Bai, and Gomes]{utsp}
Min, Y., Bai, Y., and Gomes, C.~P.
\newblock Unsupervised learning for solving the travelling salesman problem.
\newblock In \emph{NeurIPS}, 2023.

\bibitem[Nowak et~al.(2017)Nowak, Villar, Bandeira, and Bruna]{nowak}
Nowak, A.~W., Villar, S., Bandeira, A.~S., and Bruna, J.
\newblock Revised note on learning algorithms for quadratic assignment with
  graph neural networks.
\newblock \emph{arXiv preprint arXiv:1706.07450}, 2017.

\bibitem[Pan et~al.(2023)Pan, Jin, Ding, Feng, Zhao, Song, and Bian]{htsp}
Pan, X., Jin, Y., Ding, Y., Feng, M., Zhao, L., Song, L., and Bian, J.
\newblock H-{TSP}: Hierarchically solving the large-scale traveling salesman
  problem.
\newblock In \emph{AAAI}, pp.\  9345--9353, 2023.

\bibitem[Papadimitriou(1977)]{openlooptsp}
Papadimitriou, C.~H.
\newblock The euclidean travelling salesman problem is {NP}-complete.
\newblock \emph{Theoretical Computer Science}, 4\penalty0 (3):\penalty0
  237–--244, 1977.

\bibitem[Paszke et~al.(2019)Paszke, Gross, Massa, Lerer, Bradbury, Chanan,
  Killeen, Lin, Gimelshein, Antiga, et~al.]{paszke2019pytorch}
Paszke, A., Gross, S., Massa, F., Lerer, A., Bradbury, J., Chanan, G., Killeen,
  T., Lin, Z., Gimelshein, N., Antiga, L., et~al.
\newblock Pytorch: An imperative style, high-performance deep learning library.
\newblock In \emph{NeurIPS}, pp.\  8024--8035, 2019.

\bibitem[Qiu et~al.(2022)Qiu, Sun, and Yang]{dimes}
Qiu, R., Sun, Z., and Yang, Y.
\newblock {DIMES}: A differentiable meta solver for combinatorial optimization
  problems.
\newblock In \emph{NeurIPS}, pp.\  25531--25546, 2022.

\bibitem[Reeves(1993)]{localoperator}
Reeves, C.~R.
\newblock \emph{Modern heuristic techniques for combinatorial problems}.
\newblock John Wiley \& Sons, Inc., 1993.

\bibitem[Rego et~al.(2011)Rego, Gamboa, Glover, and
  Osterman]{Rego_Gamboa_Glover_Osterman_2011}
Rego, C., Gamboa, D., Glover, F., and Osterman, C.
\newblock Traveling salesman problem heuristics: Leading methods,
  implementations and latest advances.
\newblock \emph{European Journal of Operational Research}, 211\penalty0
  (3):\penalty0 427--–441, 2011.

\bibitem[Silver et~al.(2016)Silver, Huang, Maddison, Guez, Sifre, van~den
  Driessche, Schrittwieser, Antonoglou, Panneershelvam, Lanctot, Dieleman,
  Grewe, Nham, Kalchbrenner, Sutskever, Lillicrap, Leach, Kavukcuoglu, Graepel,
  and Hassabis]{Silver_mcts}
Silver, D., Huang, A., Maddison, C.~J., Guez, A., Sifre, L., van~den Driessche,
  G., Schrittwieser, J., Antonoglou, I., Panneershelvam, V., Lanctot, M.,
  Dieleman, S., Grewe, D., Nham, J., Kalchbrenner, N., Sutskever, I.,
  Lillicrap, T.~P., Leach, M., Kavukcuoglu, K., Graepel, T., and Hassabis, D.
\newblock Mastering the game of go with deep neural networks and tree search.
\newblock \emph{Nature}, 529:\penalty0 484--489, 2016.

\bibitem[Silver et~al.(2017)Silver, Schrittwieser, Simonyan, Antonoglou, Huang,
  Guez, Hubert, Baker, Lai, Bolton, Chen, Lillicrap, Hui, Sifre, van~den
  Driessche, Graepel, and Hassabis]{Silver_mcts_2}
Silver, D., Schrittwieser, J., Simonyan, K., Antonoglou, I., Huang, A., Guez,
  A., Hubert, T., Baker, L., Lai, M., Bolton, A., Chen, Y., Lillicrap, T.~P.,
  Hui, F., Sifre, L., van~den Driessche, G., Graepel, T., and Hassabis, D.
\newblock Mastering the game of go without human knowledge.
\newblock \emph{Nature}, 550\penalty0 (7676):\penalty0 354--359, 2017.

\bibitem[Sun \& Yang(2023)Sun and Yang]{difusco}
Sun, Z. and Yang, Y.
\newblock {DIFUSCO}: Graph-based diffusion solvers for combinatorial
  optimization.
\newblock In \emph{NeurIPS}, 2023.

\bibitem[Taillard \& Helsgaun(2019)Taillard and
  Helsgaun]{Taillard_Helsgaun_2019}
Taillard, E.~D. and Helsgaun, K.
\newblock {POPMUSIC} for the travelling salesman problem.
\newblock \emph{European Journal of Operational Research}, 272\penalty0
  (2):\penalty0 420--–429, 2019.

\bibitem[Vaswani et~al.(2017)Vaswani, Shazeer, Parmar, Uszkoreit, Jones, Gomez,
  Kaiser, and Polosukhin]{transformer}
Vaswani, A., Shazeer, N., Parmar, N., Uszkoreit, J., Jones, L., Gomez, A.~N.,
  Kaiser, {\L}., and Polosukhin, I.
\newblock Attention is all you need.
\newblock In \emph{NeurIPS}, pp.\  5998--6008, 2017.

\bibitem[Vinyals et~al.(2015)Vinyals, Fortunato, and Jaitly]{pointernetwork}
Vinyals, O., Fortunato, M., and Jaitly, N.
\newblock Pointer networks.
\newblock In \emph{NeurIPS}, pp.\  2692--–2700, 2015.

\bibitem[Williams(2004)]{williams1992simple}
Williams, R.~J.
\newblock Simple statistical gradient-following algorithms for connectionist
  reinforcement learning.
\newblock \emph{Machine Learning}, 8:\penalty0 229--256, 2004.

\bibitem[Wu et~al.(2021)Wu, Song, Cao, Zhang, and Lim]{wu}
Wu, Y., Song, W., Cao, Z., Zhang, J., and Lim, A.
\newblock Learning improvement heuristics for solving routing problems.
\newblock \emph{IEEE Transactions on Neural Networks and Learning Systems},
  33\penalty0 (9):\penalty0 5057--5069, 2021.

\bibitem[Ye et~al.(2024)Ye, Wang, Liang, Cao, Li, and Li]{ye2024glop}
Ye, H., Wang, J., Liang, H., Cao, Z., Li, Y., and Li, F.
\newblock {GLOP}: Learning global partition and local construction for solving
  large-scale routing problems in real-time.
\newblock In \emph{AAAI}, 2024.

\end{thebibliography}
\bibliographystyle{icml2024}

%%%%%%%%%%%%%%%%%%%%%%%%%%%%%%%%%%%%%%%%%%%%%%%%%%%%%%%%%%%%%%%%%%%%%%%%%%%%%%%
%%%%%%%%%%%%%%%%%%%%%%%%%%%%%%%%%%%%%%%%%%%%%%%%%%%%%%%%%%%%%%%%%%%%%%%%%%%%%%%
% APPENDIX
%%%%%%%%%%%%%%%%%%%%%%%%%%%%%%%%%%%%%%%%%%%%%%%%%%%%%%%%%%%%%%%%%%%%%%%%%%%%%%%
%%%%%%%%%%%%%%%%%%%%%%%%%%%%%%%%%%%%%%%%%%%%%%%%%%%%%%%%%%%%%%%%%%%%%%%%%%%%%%%
\newpage
\appendix
\onecolumn

\section{Grid Search Results for SoftDist Temperature Setting}\label{grid_search_appendix}

In our approach to fine-tuning the SoftDist temperature parameter $\tau$, we employed a two-stage grid search strategy. Initially, we conducted a coarsened grid search to broadly identify the range of effective temperatures for each TSP problem scale. This preliminary search was performed with a wide range of temperature values to quickly narrow down the potential candidates. The results of this coarsened grid search are presented in Table \ref{coarsened_grid_search}.

Following the coarsened grid search, we conducted a refined grid search within the narrowed range to find the most optimal temperature settings for each TSP scale. This second stage involved a more granular exploration of temperatures, allowing for a precise determination of the best-performing setting. The findings from this refined grid search are detailed in Table \ref{refined_grid_search}.
\begin{figure}[ht]
	\centering
	\begin{minipage}{.45\textwidth}
		\centering
		\captionof{table}{Coarsened grid search results for SoftDist temperature settings.}
		\label{coarsened_grid_search}
		\begin{center}
			\begin{small}
				\begin{sc}
					\begin{tabular}{lcc}
						\toprule
						TSP Problem & Temperature & Average Length \\
						\midrule
						\multirow{10}{*}{TSP-500} 
						& 0.0010 &  52.47558\\
						& 0.0020 &  31.59499\\
						& 0.0030 &  21.16313\\
						& 0.0040 &  17.48539\\
						& 0.0050 &  16.84009\\
						& 0.0060 &  16.78332\\
						& \textbf{0.0070} &  \textbf{16.78133}\\
						& 0.0080 &  16.78511\\
						& 0.0090 &  16.78920\\
						& 0.0100 &  16.79291\\
						\midrule
						\multirow{11}{*}{TSP-1000}
						& 0.0010 &  81.45604\\
						& 0.0020 &  38.64677\\
						& 0.0030 &  25.53236\\
						& 0.0040 &  23.71077\\
						& \textbf{0.0050} &  \textbf{23.64351}\\
						& 0.0060 &  23.64804\\
						& 0.0070 &  23.65656\\
						& 0.0080 &  23.66394\\
						& 0.0090 &  23.67698\\
						& 0.0100 &  23.69137\\
						\midrule
						\multirow{10}{*}{TSP-10000}
						& 0.0010 &  106.22613\\
						& \textbf{0.0020} &  \textbf{74.10114}\\
						& 0.0030 &  74.24206\\
						& 0.0040 &  74.48813\\
						& 0.0050 &  74.77912\\
						& 0.0060 &  75.08760\\
						& 0.0070 &  75.43125\\
						& 0.0080 &  75.73975\\
						& 0.0090 &  76.09572\\
						& 0.0100 &  76.45497\\
						\bottomrule
					\end{tabular}
				\end{sc}
			\end{small}
		\end{center}
	\end{minipage}%
	\hfill
	\begin{minipage}{.45\textwidth}
		\centering
		\captionof{table}{Refined grid search results for SoftDist temperature settings.}
		\label{refined_grid_search}
		\begin{center}
			\begin{small}
				\begin{sc}
					\begin{tabular}{lcc}
						\toprule
						TSP Problem & Temperature & Average Length \\
						\midrule
						\multirow{10}{*}{TSP-500} 
						& 0.0060 & 16.78332 \\
						& 0.0061 & 16.78390 \\
						& 0.0062 & 16.78535 \\
						& 0.0063 & 16.78268 \\
						& 0.0064 & 16.78538 \\
						& 0.0065 & 16.78185 \\
						& \textbf{0.0066} & \textbf{16.78020} \\
						& 0.0067 & 16.78195 \\
						& 0.0068 & 16.78463 \\
						& 0.0069 & 16.78320 \\
						\midrule
						\multirow{11}{*}{TSP-1000}
						& 0.0050 & 23.64351 \\
						& \textbf{0.0051} & \textbf{23.63891} \\
						& 0.0052 & 23.64239 \\
						& 0.0053 & 23.64302 \\
						& 0.0054 & 23.64231 \\
						& 0.0055 & 23.64560 \\
						& 0.0056 & 23.64476 \\
						& 0.0057 & 23.64931 \\
						& 0.0058 & 23.64592 \\
						& 0.0059 & 23.64683 \\
						\midrule
						\multirow{10}{*}{TSP-10000}
						& 0.0010 & 106.22613 \\
						& 0.0011 & 91.68151 \\
						& 0.0012 & 83.10377 \\
						& 0.0013 & 78.16867 \\
						& 0.0014 & 75.74385 \\
						& 0.0015 & 74.72504 \\
						& 0.0016 & 74.26747 \\
						& 0.0017 & 74.13749 \\
						& \textbf{0.0018} & \textbf{74.07734} \\
						& 0.0019 & 74.09550 \\
						\bottomrule
					\end{tabular}
				\end{sc}
			\end{small}
		\end{center}
	\end{minipage}
\end{figure}

\section{Performance Analysis under Varying Time Budgets with UTSP's MCTS Settings}\label{appendix:utsp_results}

We present the performance of MCTS under the settings proposed by \citet{utsp}, offering a supplementary perspective to our main experiments. Our findings indicate that the performance under UTSP's MCTS settings closely mirrors that observed with the default MCTS settings, with LKH-3 consistently outperforming MCTS across the experiments. Additionally, the performance of various methods, including those using a zero-input heatmap, perform similarly, indicating the limited influence of the heatmap on the MCTS with \citet{utsp}'s settings.
\begin{figure}[ht]
	\vskip 0.2in
	\centering
	\begin{subfigure}[b]{0.45\textwidth}
		\includegraphics[width=\textwidth]{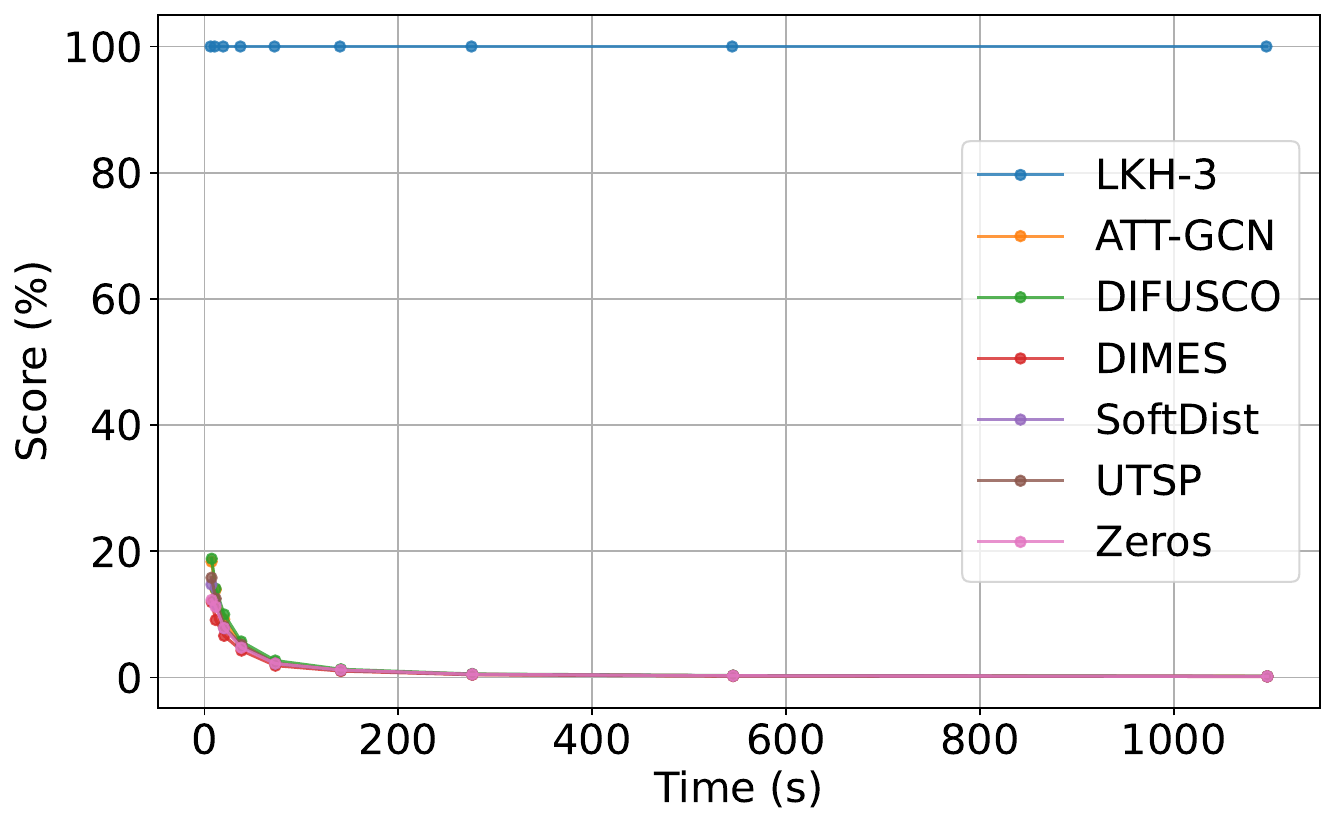}
		\caption{TSP-500 with different MCTS settings.}
		\label{fig:tsp500_utsp}
	\end{subfigure}
	\hfill
	\begin{subfigure}[b]{0.45\textwidth}
		\includegraphics[width=\textwidth]{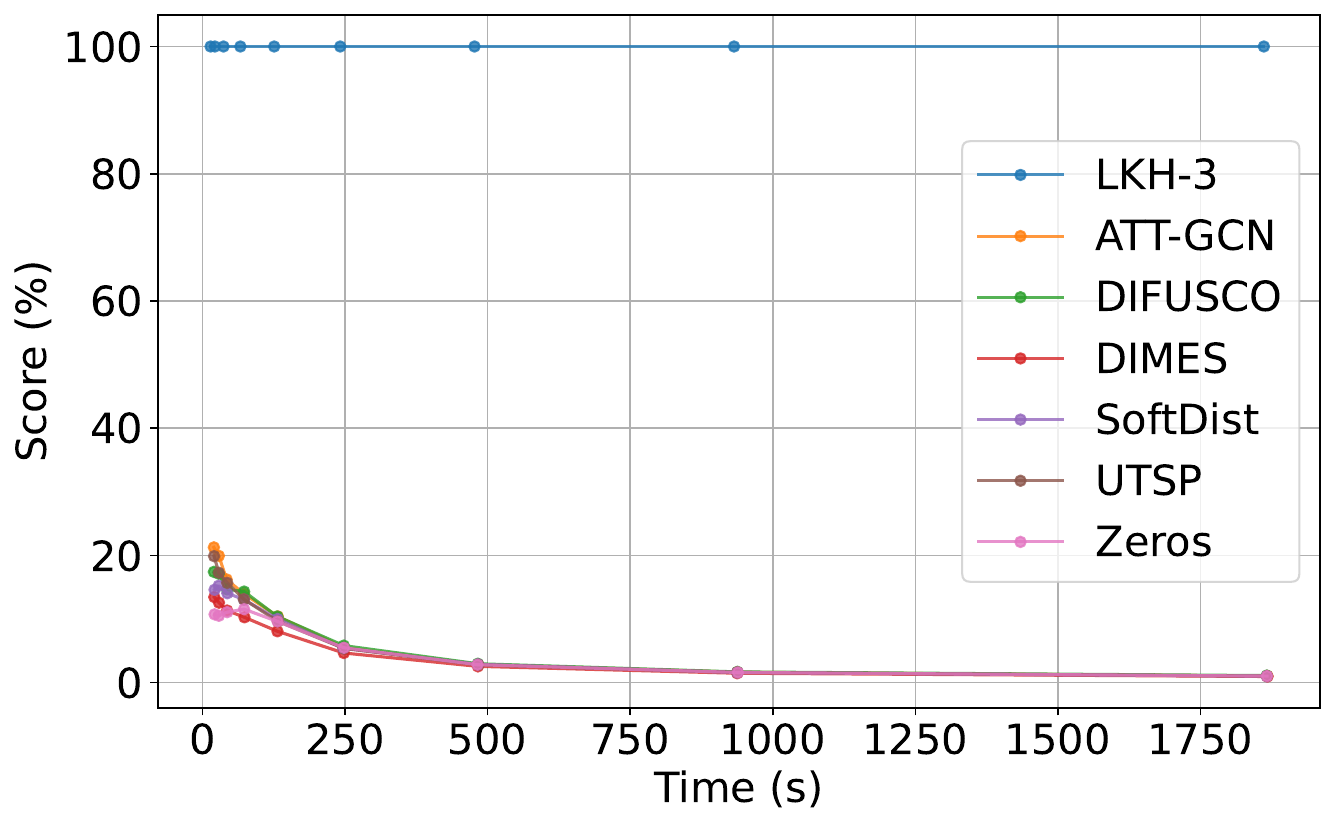}
		\caption{TSP-1000 with different MCTS settings.}
		\label{fig:tsp1000_utsp}
	\end{subfigure}
	\caption{Performance of MCTS under different settings by \cite{utsp} for TSP-500 and TSP-1000.}
	\label{fig:tsp_utsp}
\end{figure}

\section{Runnable PyTorch Code for SoftDist-Based Heatmap Generation}

We have provided a directly runnable Python code, implemented using PyTorch \cite{paszke2019pytorch}. The input \texttt{batch\_coords} represents a batch of TSP-$n$ problems, where each problem is represented as $n$ two-dimensional coordinates, forming a tensor of size $(batch\_size, n, 2)$. The \texttt{tau} parameter is the temperature $\tau$ in Equation \ref{softdist}, determined through grid search. This function outputs a tensor of size $(batch\_size, n, n)$, generating a corresponding heatmap for each TSP problem.

\begin{lstlisting}[language=Python]
import torch
import torch.nn.functional as F


def create_heatmap_matrix(batch_coords, tau, device="cuda:0"):
batch_coords = torch.tensor(batch_coords, device=device).float()

coord_diff = batch_coords[:, :, None, :] - batch_coords[:, None, :, :]

distance_matrix = torch.sqrt(torch.sum(coord_diff ** 2, dim=-1))

eye = torch.eye(distance_matrix.size(1), device=device).unsqueeze(0)
distance_matrix = torch.where(
eye == 1, 
torch.tensor(float('inf'), dtype=torch.float, device=device), 
distance_matrix
)

heatmap = F.softmax(-distance_matrix / tau, dim=2)

return heatmap.cpu().numpy()

\end{lstlisting}

\section{Extended Related Work}
Existing methods to tackle the TSP fall into two broad categories: machine learning-based and non-learning algorithms. In this section, we focus exclusively on machine learning-based approaches. For non-learning algorithms, interested readers are directed to \citet{Applegat2007, Helsgaun_2009, Rego_Gamboa_Glover_Osterman_2011, Helsgaun_2017, Taillard_Helsgaun_2019} for further exploration.

Machine learning-based approaches for solving TSP can be broadly classified into two categories, based on the method of solution construction. The first category, construction-based methods, progressively builds a solution by sequentially adding new points to an incomplete path in an autoregressive manner until a complete path is formed. The second category, search-based methods, starts with a complete solution and continuously applies local OR operations \cite{localoperator} in an effort to improve it. This classification reflects a fundamental divide in strategy: while construction-based methods focus on incrementally creating a route, search-based methods revolve around refining an already established route.

\subsection{Construction-based Methods}

Construction-based methods in machine learning for solving the TSP have evolved significantly over the years. Early approaches like the Pointer Network (PointerNet) \cite{pointernetwork} proposed an end-to-end approach that decodes TSP solutions autoregressively from scratch using recurrent neural networks. However, this supervised learning method requires a large number of pre-computed optimal (at least high-quality) TSP solutions, being unaffordable for large-scale instances. This framework was further enhanced by integrating reinforcement learning for better performance and generalization, as seen in the work of \citet{bello}, where the negative tour length serves as a reward signal to guide an actor-critic architecture. The emergence of Transformer architectures \cite{transformer}, known for their success in the text generation domain, has further revolutionized this field \cite{kool,pomo,sym-nco} by supplanting PointerNet. These methods, while effective for smaller TSP instances up to about 100 nodes, encounter scalability challenges and latency in inference when dealing with larger numbers of cities \cite{rethink,attgcn}. This is due to the action space growing linearly and the quadratic complexity inherent in the self-attention mechanism \cite{transformer}.

One exception is \citet{htsp}, which employs a hierarchical divide-and-conquer strategy by decomposing a large-scale TSP problem into smaller open-loop TSP sub-problems \cite{openlooptsp}. While this hierarchical approach reduces training complexity, enabling scalability to large instances (e.g., up to 10,000 points), it trades off solution quality: the partitioning strategy limits the solution quality, resulting in a notable reduction in performance. For instance, the optimality gap reaches 6.62\% for TSP-1000 and 7.32\% for TSP-10000.

\subsection{Search-based Methods}

In contrast to construction-based methods, search-based methods aim to improve existing solutions through iterative refinement until computational budgets are exhausted. These methods rely on classical local operators, such as local search by \citet{chen_tian,lu}, ruin-and-repair by \citet{hottung} and 2-opt by \citet{costa,dual,wu}. However, improvement heuristic learners encounter the sparse reward problem when dealing with large graphs \cite{rethink,bengio}, and overly simplistic local operators can limit the performance of the algorithms. One variant is \citet{ye2024glop}, which adopts a divide-and-conquer strategy and utilizes search-based methods for improving the smaller subproblems. However, similar to \citet{htsp}, it also trades off solution quality to reduce training complexity.

Recent successes in addressing large-scale TSP problems \cite{attgcn,dimes,difusco,utsp} have utilized Monte Carlo tree search \cite{Coulom_mcts,Silver_mcts,Silver_mcts_2} as a powerful post-processing algorithm. These emerging algorithms can be divided into two stages: heatmap generation and MCTS with guidance from the heatmap. Initially, deep learning models are trained to generate heatmaps, providing scores for each edge's selection (which indicates the probability of each edge belonging to the optimal solution) as mentioned in \citet{nowak,joshi,realworld,dpdp}, where a specific surrogate loss function is designed by supervised learning \cite{attgcn,difusco}, or unsupervised learning \cite{utsp}, or reinforcement learning \cite{dimes}. Subsequently, these heatmaps will act as priors to guide the MCTS. This heatmap-guided MCTS method has achieved satisfactory results in solving large-scale TSP problems, reaching state-of-the-art performance. For a visual representation of this method, please refer to Figure \ref{fig:heatmap_guided_mcts}.

\end{document}